\documentclass[11pt]{article}

\usepackage{ACL2023}
\usepackage{algorithm}  
\usepackage{algpseudocode}  
\usepackage{amsmath}

\usepackage{times}
\usepackage{latexsym}
\usepackage{graphicx}
\usepackage{booktabs}

\usepackage[T1]{fontenc}

\usepackage[utf8]{inputenc}

\usepackage{microtype}
\usepackage{amsfonts}

\title{Attack Named Entity Recognition by Entity Boundary Interference}

\author{Yifei Yang$^{1,2,\dag}$, Hongqiu Wu$^{1,2,\dag}$ and Hai Zhao$^{1,2,}$\thanks{$\ $ Corresponding author. $^\dag$ Equal contribution.}\\
$^{1}$Department of Computer Science and Engineering, Shanghai Jiao Tong University\\
$^{2}$MoE Key Lab of Artificial Intelligence, AI Institute, Shanghai Jiao Tong University\\
{\tt \{yifeiyang, wuhongqiu\}@sjtu.edu.cn, zhaohai@cs.sjtu.edu.cn}}

\begin{document}
\maketitle
\begin{abstract}

Named Entity Recognition (NER) is a cornerstone NLP task while its robustness has been given little attention. This paper rethinks the principles of NER attacks derived from sentence classification, as they can easily violate the label consistency between the original and adversarial NER examples. This is due to the fine-grained nature of NER, as even minor word changes in the sentence can result in the emergence or mutation of any entities, resulting in invalid adversarial examples. To this end, we propose a novel one-word modification NER attack based on a key insight, NER models are always vulnerable to the boundary position of an entity to make their decision. We thus strategically insert a new boundary into the sentence and trigger the \textit{Entity Boundary Interference} that the victim model makes the wrong prediction either on this boundary word or on other words in the sentence. We call this attack \textit{Virtual Boundary Attack (ViBA)}, which is shown to be remarkably effective when attacking both English and Chinese models with a 70\%-90\% attack success rate on state-of-the-art language models (e.g. RoBERTa, DeBERTa) and also significantly faster than previous methods.


\end{abstract}

\section{Introduction}

The goal of Named Entity Recognition (NER) is to find the predefined named entities such as locations, persons, and organizations in a sentence. It is a fundamental task in natural language processing (NLP) behind various downstream applications  \cite{clark2018neural,sil2013re,babych2003improving,nikoulina2012hybrid}. 

\begin{figure}[!t]
    \centering
    \includegraphics[width=\linewidth,scale=1]{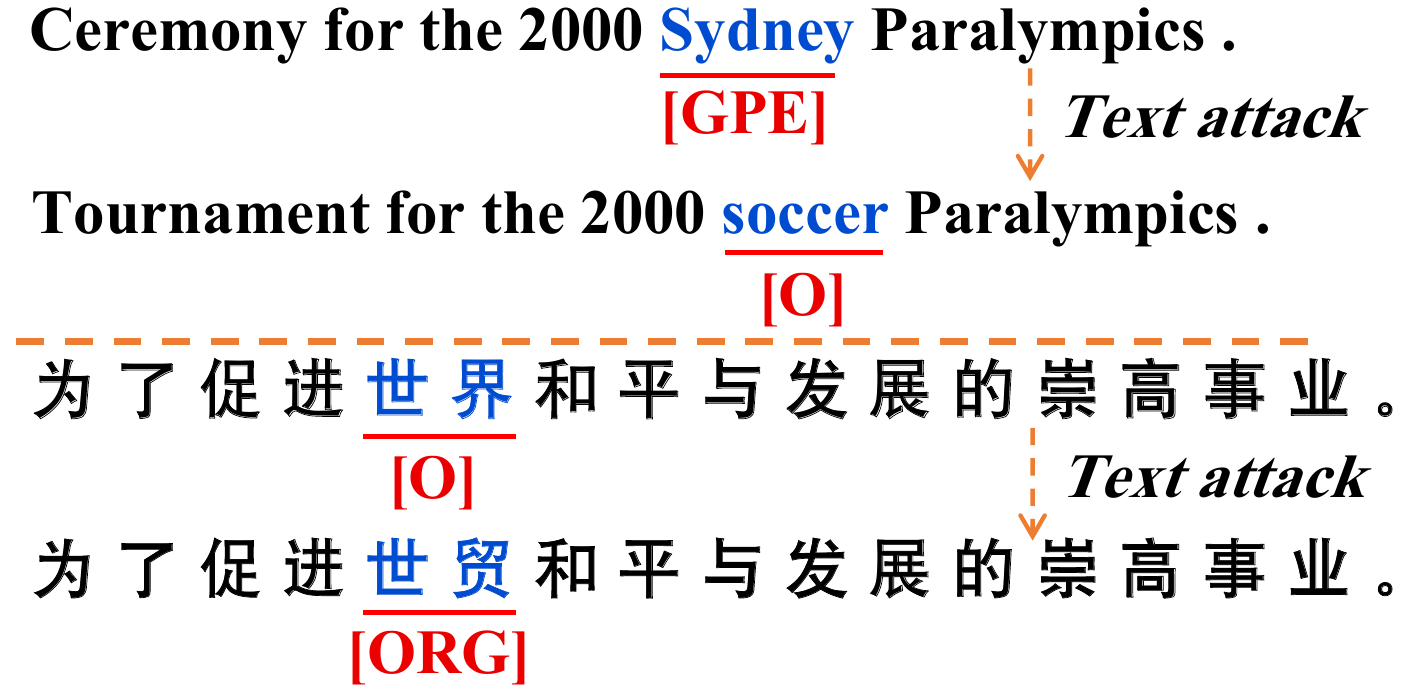}
    \caption{A case of Label Shift.}
    \label{fig:shift}
\end{figure}


Language models have been shown to be vulnerable to cunningly crafted input data, producing misjudgments, thereby undermining their security and trustworthiness.
Great attention has been paid to study the robustness of language models to reveal its vulnerabilities and deficiencies, for the sake of providing defense techniques, e.g. on sentence classification, question answering.
However, the study on the robustness of NER models is still lacking.   


Recently, \citeauthor{simoncini2021seqattack} first pay attention to attacking NER models and try to adapt the text attack methods on sentence classification to NER. \citeauthor{lin-etal-2021-rockner} propose an adversarially generated NER dataset RockNER, by word substitution also following the procedure of sentence classification attackers.

However, we find that the conventional principles of text attack on sentence classification can easily violate the label consistency between original and adversarial NER examples.
Specifically, the current attackers apply word insertion, swapping or substitution while maintaining the semantic similarity to keep the sentence label unchanged as possible.
As opposed to sentence classification, NER is a finer-grained structure prediction task, where any word changes like insertion, swapping, and substitution, can result in the emergence of new entities or mutation of original entities.


We call this notion \textit{label shift}. We show a case in Figure \ref{fig:shift}, where a GPE (geopolitical) entity \textit{Sydney} in the original sentence is replaced by \textit{soccer}, and \textit{world} is replaced by \textit{WTO} by the attacker \cite{morris2020textattack}. However, \textit{soccer} obviously cannot be a GPE and \textit{WTO} should be an ORG, incurring an invalid adversarial example. We find such an issue widely existed in current attackers, which has a significant negative impact.

The fine-grained nature of NER determines that one should make as few modification to the sentence as possible in order not to produce label shift.
Thus, this paper proposes a novel NER attack, which only modifies one word in the original sentence to alleviate label shift.
Moreover, the impact of label shift can be perfectly avoided due to ingenious technique design when judging the validity of a candidate example. Our method is based on a key insight, that NER models are always sensitive to the boundary word of an entity.
Specifically, when inserting a boundary token (i.e. the first/last token of an entity) into a sentence, the nowadays NER models will be easily fooled and exhibit some abnormal behaviors. We call such a phenomenon as \textit{Entity Boundary Interference} (EBI).

The contributions of this paper are below:

$\bullet$ We propose \textit{Virtual Boundary Attack} (ViBA) based on the EBI, an novel NER attacker, which avoids the label shift problem that other attackers suffer from. We evaluate ViBA on several state-of-the-art pre-trained language models on widely used English and Chinese benchmarks and experiments show that ViBA has a high attack success rate. It also has a good efficiency advantage with almost a linear time complexity.

$\bullet$ We conduct in-depth analysis on the causes of EBI and interpret the effectiveness of ViBA.

$\bullet$ We propose two defending strategies to train NER models against ViBA.

\begin{figure}[!t]
    \centering
    \includegraphics[width=\linewidth,scale=1.00]{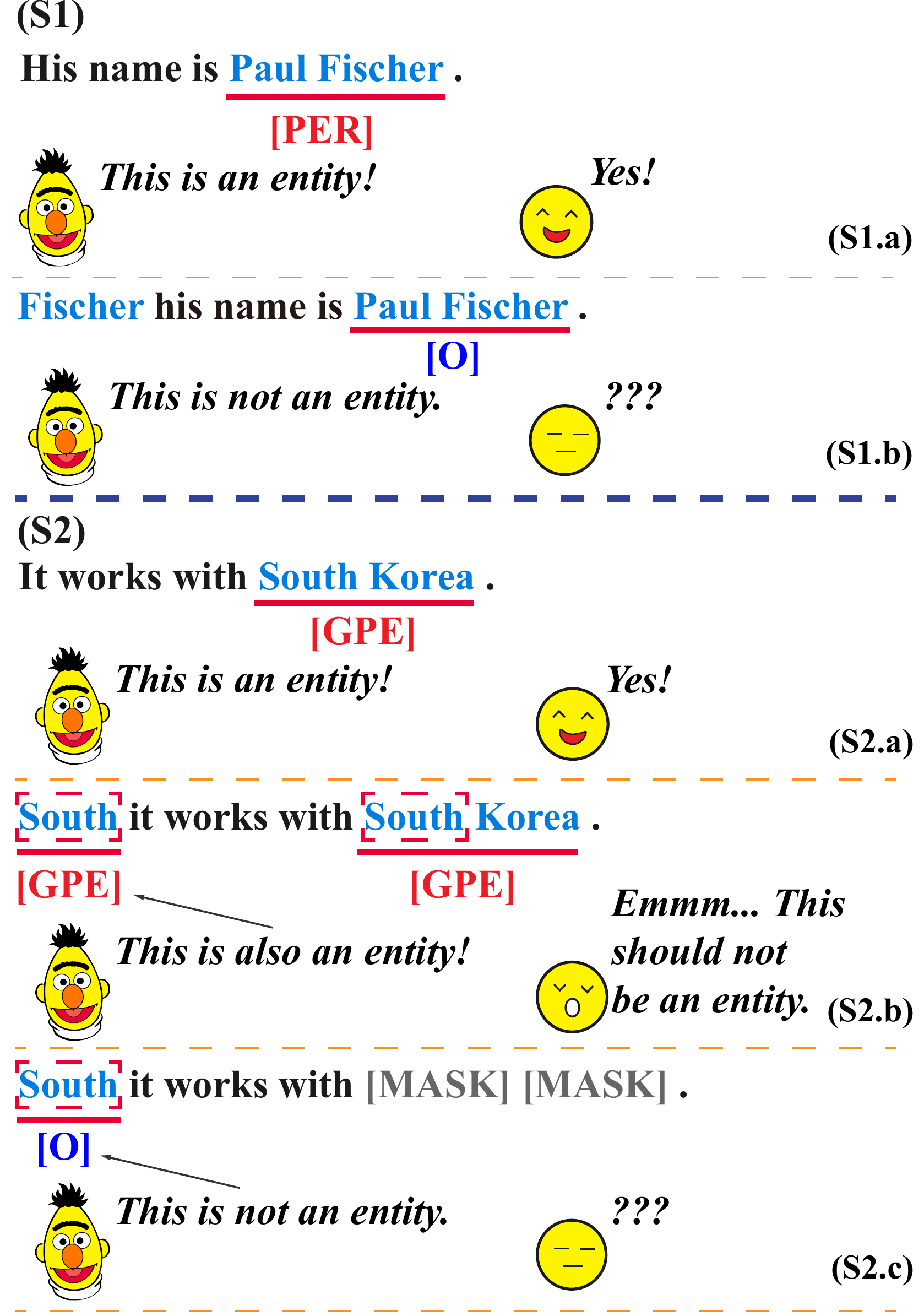}
    \caption{A demonstration of Entity Boundary Interference.}
    \label{fig:ViBA_example}
\end{figure}

\section{Method}
\subsection{Entity Boundary Interference}
Previous studies assume that an NER model is heavily reliant on the boundary of an entity \cite{peng2016improving, tan2020boundary} when making decisions. Given an entity, the boundary refer to its leftmost or rightmost token.
In the light of this assumption, our vision is that NER models can be vulnerable if the attacker attempt to manipulate these boundary tokens.
Figure \ref{fig:ViBA_example} demonstrates two representative phenomenons where the model falls into mistakes when there is a new boundary inserted in the sentence at some positions.

$\bullet$ \textbf{S1:} Insertion of semantically unrelated boundary may changes the predictions of other entities. As shown in Figure \ref{fig:ViBA_example} (S1), the model correctly recognizes \textit{Paul Fischer} as a PER (Person) entity in (S1.a). When we insert the right boundary \textit{Fischer} at the beginning of the sentence in (S1.b), surprisingly, the model no longer recognizes \textit{Paul Fischer} as a PER, even if it still is. Apparently, human will not make such a mistake.

$\bullet$ \textbf{S2:} The model may mistakenly assume a correlation between the inserted boundary and the original entity. In Figure \ref{fig:ViBA_example} (S2), the model first wrongly recognizes the inserted \textit{South} as a GPE (geopolitical) in (S2.b). Paradoxically, it is no more after the original entity \textit{South Korea} is masked in (S2.c). It indicates that the model pathologically assumes the co-occurring boundaries are relevant, which is different from the way humans perceive text and should be regarded as another non-robust phenomenon.

S1 and S2 show that there is a coupling effect between the model recognition of different entities in the sentence.
In S1, the emergence of a new entity \textit{Fischer} causes a flip in the prediction of \textit{Paul Fischer}.
In S2, the erasure (being masked) of an original entity \textit{South Korea} causes a miss recall of another entity \textit{South}. The underlying is that the prediction of \textit{South} is coupled with the co-occurrence of \textit{South Korea}.
We notice that these entities are supposed not to have any connection. We call the above phenomenon \textit{Entity Boundary Interference} (EBI).

\subsection{ViBA}

We introduce \textit{Virtual Boundary Attack} (ViBA), a novel attack algorithm for NER models based on our finding of EBI.
ViBA attacks the model by inserting a boundary token of some entity into the sentence. The goal is to induce wrong predictions of the model as in S1 and S2.
We denote the inserted boundary ``virtual boundary'' for the reason that the inserted boundary is not a real entity.
Algorithm \ref{algor:ViBA} summarizes the procedure of ViBA:

\noindent \textbf{(1) Predict (line 1-3).}

Given an input sentence $\mathcal{X}=\mathbf{x}_1,\mathbf{x}_2,\cdots,\mathbf{x}_n$, we first feed it to the victim model to obtain the original prediction $\mathcal{Y}$, which is a list of predicted named entity tags and has the same length with $\mathcal{X}$. Each tag in $\mathcal{Y}$ is a pre-defined abbreviated label such as ``PER'' (Person), ``LOC'' (Location), etc. Following the convention, ``O'' refers to a non-entity token. Then we extract the named entities $\mathcal{E}$ as well as their corresponding positions $\mathcal{L}$ from $\mathcal{Y}$.

\noindent \textbf{(2) Decide Safety Areas (line 4).}

\begin{figure}[!tp]
    \centering
    \includegraphics[width=\linewidth,scale=1.00]{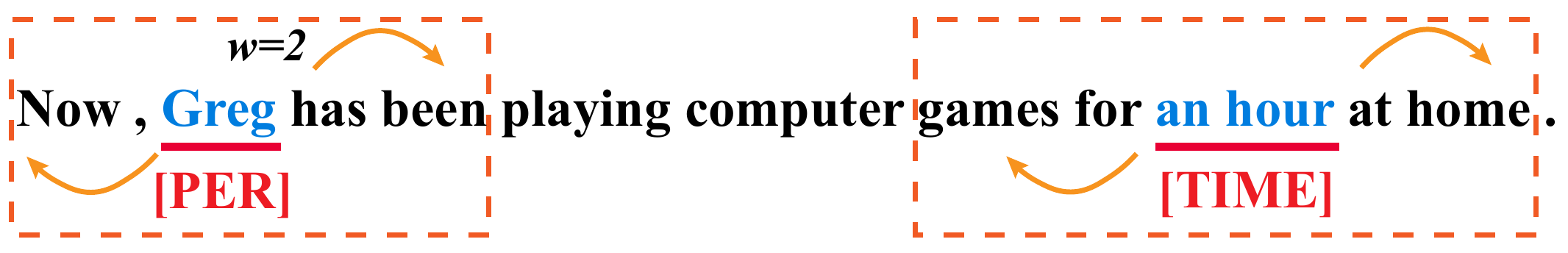}
    \caption{A case of safety areas.}
    \label{fig:safe}
\end{figure}

We set safety areas to prevent two risky operations based on a safety distance $w$. For one thing, it is not allowed to insert a boundary inside an entity because it would undermine the entity and trigger the label shift problem. For another thing, an entity tag is likely to mutate when its local context changes. For example, the inserted boundary may form a new entity with its surrounding tokens. We show a case of safety areas in Figure \ref{fig:safe}.

\noindent \textbf{(3) Attack (line 5-9).}

We next generate the candidate adversarial examples. For each named entity $e$ in $\mathcal{E}$, we extract its leftmost and rightmost boundaries. Then, we go through every position in the sentence outside the safety areas and insert the boundary to generate a candidate example $\mathcal{X}^{\prime}$. Then $\mathcal{X}^{\prime}$ is fed into the victim model to obtain its prediction $\mathcal Y^{\prime}$.

\begin{table*}[!t]
\centering
{\begin{tabular}{l|cccc}
\toprule

\textbf{Test set} & \textbf{WNUT}   & \textbf{OntoNotes-en} & \textbf{MSRA} & \textbf{OntoNotes-ch} \\ \hline
\begin{tabular}[c]{@{}l@{}}\textbf{Samples} \end{tabular}  & 686 / 1287 & 4561 / 9479 & 2344 / 4365  & 2392 / 4472  \\ \hline
\begin{tabular}[c]{@{}l@{}}\textbf{Entities per sample} \end{tabular}  & 1.57 & 2.45 & 2.61 & 3.13 \\ \hline
\begin{tabular}[c]{@{}l@{}}\textbf{Tokens per sample} \end{tabular} & 19.67 & 24.08 & 47.3 & 45.06  \\ \hline 

\end{tabular}}
\caption{Statistics for each used test set.}
\label{table:static}
\end{table*}

\noindent \textbf{(4) Check Success (line 10-17).}

The following two criteria are applied to determine whether the attack is successful:

\textbf{\textit{Criterion 1 (line 10-12).}} This criterion corresponds to the EBI S1, that the inserted token should not affect the predictions of the original entities. Specifically, we check the consistency of $\mathcal Y$ and $\mathcal Y^{\prime}$. Note that we set a safety area for the inserted position during the comparison in order to avoid the label shift condition that the inserted boundary is an entity or forms a new entity with surrounding tokens. Any inconsistency indicates a successful attack due to the EBI S1.

\begin{algorithm}[!tb]
\caption{Virtual Boundary Attack}
\label{algor:ViBA}
\begin{algorithmic}[1]
    \Require  
        Victim model $\mathcal{F}$, input example $\mathcal X$, safety distance $w$.
    \Ensure Adversarial example $\mathfrak X$.
    \State $\mathcal Y\gets\mathcal{F}(\mathcal X)$
    \State $\mathcal E\gets$ Extract each entity in $\mathcal X$ following $\mathcal Y$
    \State $\mathcal L\gets$ Locate each entity in $\mathcal X$ following $\mathcal Y$
    \State $\mathcal S\gets$ Decide safety area following $\mathcal L$ and $w$
    \For{$e$ in $\mathcal E$}
        \For{$j$ in $\{1 \sim n\} \setminus \mathcal S$}
            \For{$b$ in $\{e^{left},e^{right}\}$}
                \State $\mathcal X^{\prime}\gets$ Replace $\mathcal X_{[j]}$ with $b$
                                
                \State $\mathcal Y^{\prime}\gets\mathcal{F}(\mathcal X^{\prime})$

                \If{$\mathcal Y^{\prime} \setminus \mathcal Y^{\prime}_{[j-w:j+w+1]} \neq \mathcal Y$}
                    \State \Return $\mathcal X^{\prime}$
                \EndIf                
                \State $\mathcal X^{\prime}_{m}\gets$ Mask $e$ in $\mathcal X^{\prime}$
                \State $\mathcal Y^{\prime}_{m}\gets\mathcal{F}(\mathcal X^{\prime}_{m})$
                
                \If{$\mathcal Y^{\prime}_{[j]} \neq \mathcal Y^{\prime}_{m[j]}$}
                    \State \Return $\mathcal X^{\prime}$
                \EndIf

            \EndFor
        \EndFor
    

    
    \EndFor
    \State \Return None

\end{algorithmic}
\end{algorithm}

\textbf{\textit{Criterion 2 (line 13-17).}} This criterion corresponds to the EBI S2, that the model prediction of the virtual boundary should not change after we mask the original entity. We replace the named entity in the $\mathcal{X}^{\prime}$ with \texttt{[MASK]} and get $\mathcal{X}^{\prime}_{m}$. Then, the prediction $\mathcal Y^{\prime}_{m}$ of $\mathcal{X}^{\prime}_{m}$ is given by the victim model. Inconsistent prediction represents the S2 occurs.

It is worth mentioning that ViBA perfectly avoids the label shift by the following properties:

$\bullet$ The safety areas guarantee that the original entity tags will not be affected after the insertion of a new boundary.

$\bullet$ Criterion 2 is independent of labels since we only care about the prediction of the virtual boundary.

\section{Experiments}

\begin{table*}[t]
\small
\normalsize
\centering
\setlength\tabcolsep{3.5pt}
\begin{tabular}{lcccc|cccc}
\toprule
                               & \multicolumn{4}{c|}{\textit{\textbf{English}}}                              & \multicolumn{4}{c}{\textit{\textbf{Chinese}}}                              \\ \hline \hline
                               & \multicolumn{2}{c}{\textbf{WNUT}} & \multicolumn{2}{c|}{\textbf{OntoNotes-en}} & \multicolumn{2}{c}{\textbf{MSRA}} & \multicolumn{2}{c}{\textbf{OntoNotes-ch}} \\ \hline
                               & \textbf{ASR}     & \textbf{SS}    & \textbf{ASR}        & \textbf{SS}       & \textbf{ASR}     & \textbf{SS}    & \textbf{ASR}       & \textbf{SS}       \\ \hline
\textbf{BERT}$_{\rm base}$     & 57.1/59.6             & 98.0/95.4           & 73.2/\textbf{75.1}       & 98.1/96.5              & 91.2/91.4             & 98.8/98.4           & 85.5/86.4               & 98.7/98.2              \\ \hline
\textbf{RoBERTa}$_{\rm large}$ & 67.1/\textbf{67.8}    & 97.9/95.5          & 70.0/73.0               & 98.1/96.4              & 91.7/92.3            & 98.8/98.3           & 86.9/89.1               & 98.1/98.2              \\ \hline
\textbf{DeBERTa}$_{\rm large}$ & 56.1/62.5             & 98.0/95.7           & 70.7/74.7                & 98.1/96.4              & -                & -              & -                  & -                 \\ \hline
\textbf{MacBERT}$_{\rm large}$ & -                & -              & -                   & -                 & \textbf{93.2}/92.0    & 98.8/98.3           & 89.4/\textbf{89.8}      & 98.6/98.1             \\ \bottomrule
\end{tabular}
\caption{The attack success rate and semantic similarity for ViBA/ViBA-rep.}
\label{table:main_res}
\end{table*}

\begin{table}[!ht]

\centering
\setlength\tabcolsep{0.5pt}
\small
\begin{tabular}{lcccc}
\toprule
   & \multicolumn{1}{c}{\textbf{WNUT}} & \multicolumn{1}{c}{\textbf{OntoNotes-en}} & \multicolumn{1}{c}{\textbf{MSRA}} & \multicolumn{1}{c}{\textbf{OntoNotes-ch}} \\ \hline
                             
\textbf{BERT}    & 54.9/75.7 & 22.3/55.4  & 42.6/75.8 & 55.4/61.8   \\ \hline
\textbf{RoBERTa} & 41.0/67.9    & 17.1/52.6   & 37.1/76.8               & 56.6/65.7      \\ \hline
\textbf{DeBERTa} & 42.6/73.6. & 14.5/51.0. & -  & - \\ \hline
\textbf{MacBERT} & -  & -   & 46.5/77.4  & 60.5/71.8 \\ \bottomrule
\end{tabular}
\caption{EASR1/EASR2 for ViBA on different datasets towards four victim models.}
\label{table:EASR}
\end{table}

\subsection{Datasets}


\quad $\bullet$ \textbf{OntoNotes5.0} \cite{weischedel2013ontonotes} is a multilingual NER dataset of Chinese, English and Arabic. There are eighteen types of named entities, eleven of which are types like Person, Organization and seven are values such as Date, Percent. In this paper, we select the popular Chinese and English versions for our experiments.

$\bullet$ \textbf{MSRA} \cite{levow2006third} is one of the commonly used Chinese NER datasets which accommodates three named entity types and the data in MSRA are collected from the news domain.


$\bullet$ \textbf{WNUT2017} \cite{derczynski-etal-2017-results} is an English NER dataset which has six types. It focuses on identifying unusual, previously-unseen entities and is more challenging.

These benchmarks have standard train/dev/test split. Some statistical data of the test sets are shown in Table \ref{table:static}. The total number of sentences containing at least one entity / the sizes of datasets are shown in the Samples row. We also count the average amount of entities in each sentence and the average sentence length.



\subsection{Metric}


\quad $\bullet$ \textbf{Attack Success Rate (ASR)} is the main measurement of the attacker's effectiveness towards a victim model, which is the ratio of the achieved successful examples over all examples. A higher ASR suggests a more effective attacker.

$\bullet$ \textbf{Semantic Similarity (SS)} serves as a semantic measurement between two sentences. We leverage \textit{text2vec} for evaluation \cite{Text2vec}. A greater SS leads to a better adversarial example.

$\bullet$ \textbf{Entity-Level Attack Success Rate (EASR)} is a ViBA-specific metric to imply how frequently the EBI occurs, which is the proportion of entities that can successfully trigger the EBI out of all entities. EASR1 and EASR2 are corresponding to S1 and S2, respectively.

$\bullet$ \textbf{Edit Distance (ED)} reflects the syntax similarity between two sentences. We except to attack a model by an example that maintains high syntax similarity to the original one.




\begin{table*}[!tp]
\centering
\begin{tabular}{lcccc}
\toprule
   & \multicolumn{1}{c}{\textbf{WNUT}} & \multicolumn{1}{c}{\textbf{OntoNotes-en}} & \multicolumn{1}{c}{\textbf{MSRA}} & \multicolumn{1}{c}{\textbf{OntoNotes-ch}} \\ \hline
                             
    \textbf{RockNER}    & 64.3/90.3/2.2 & 17.1/84.1/2.1  & 53.6/94.8/2.3 & 72.2/95.0/2.3   \\
    \textbf{CLARE}    & 55.5/95.4/1.2 & 55.0/95.9/1.2 & 56.4/94.9/5.9 & 40.7/96.3/2.6   \\
    \hline
    \textbf{ViBA} & \textbf{67.1}/\textbf{97.9}/\textbf{1.0} & \textbf{70.0}/\textbf{98.1}/\textbf{1.0} & \textbf{91.7}/\textbf{98.8}/\textbf{1.0} & \textbf{86.9}/\textbf{98.1}/\textbf{1.0} \\
 \bottomrule
\end{tabular}
\caption{ASR$\uparrow$/SS$\uparrow$/ED$\downarrow$ comparisons of ViBA and state-of-the-art attackers.}
\label{tab:campara_ViBA_others}
\end{table*}

\subsection{Settings}
We evaluate ViBA on the BERT-base \cite{devlin2018bert}, RoBERTa-large \cite{liu2019roberta} models of Chinese and English versions. In addition, DeBERTa-large \cite{he2020deberta} is leveraged for the evaluation of the English datasets. MacBERT-large \cite{cui-etal-2020-revisiting} is used for the Chinese datasets. We first fine-tune the models on the training set for 6 epochs and select the best trained checkpoints by dev set. Then we apply ViBA to attack them on the test set. We have heuristically set the safety distance and detect $w=2$ can effectively prevent the inserted boundary from forming new entities with surrounding words for all experiments. We conduct experiments on a single NVIDIA RTX 3090 GPU.

\subsection{Main Results}


We evaluate ViBA for multiple models on different Chinese and English datasets, and the results are shown in Table \ref{table:main_res}. Considering that the insertion will change the length of the sentence and cause too obvious a distinction, we also change the ``insert'' operation in ViBA to the ``replace'' operation for comparison, named as ViBA-rep. Overall, ViBA achieves high ASR when attacking both Chinese and English datasets. The ASR on the Chinese datasets is as high as 85\% - 93\%. Although relatively lower on the English datasets, the ASR is ranging from 55\% to 73\% which is still an ideal performance. It is noteworthy that the English datasets generally have shorter sentences and less entities, their smaller search spaces will lead to relatively poorer ASR. Comprehensively, ViBA is still a great attacker on the above benchmarks.

In Table \ref{table:main_res}, the average SS between the adversarial and original examples of all datasets exceeds 97.9, which guarantees that (1) the semantics of the adversarial examples are extremely close to the original ones and (2) the adversarial examples are natural and look similar to the original ones. 

Generally, ViBA-rep exhibits higher ASR than vanilla ViBA. But replacement fails to retain all the tokens and generates examples with a greater semantic difference, as its lower SS. Considering ASR and SS comprehensively, we conduct follow-up experiments all on vanilla ViBA.

To explore occurrence frequency of S1 and S2, we present in Table \ref{table:EASR} the EASR1 and EASR2 of ViBA. Since many entities can induce both S1 and S2, the sum of EASR1 and EASR2 may exceed 1.0. It can be seen from that (1) both EASR1 and EASR2 are generally at a high level, indicating that S1 and S2 are both frequent non-robust phenomena so the NER models are fragile to the boundary tokens and (2) S2 occurs more frequently than S1.

To compare ViBA with other state-of-the-art attackers, we reproduce the RockNER \cite{lin-etal-2021-rockner} and CLARE \cite{li2021contextualized} adapted for NER on the four datasets, where the victim model is RoBERTa-large. When adapting the previous algorithms, we retain their processes of perturbing sentence but change the success judgement to whether the predicted tag sequences have changed. We compare the ASR/SS/ED of different attackers in Table \ref{tab:campara_ViBA_others}. For ASR, it is displayed that ViBA easily outperforms the previous attackers. Since the previous methods will trigger the label shift problem, their actual ASR should be lower than the reported value, which further highlights our effectiveness. Better SS proves that ViBA preserves more semantic similarity. It is worth mentioning that ViBA is an one-word modification attacker and always maintains the ED to 1.0, which shows that it keeps better syntax than all the previous attackers. Both advantageous SS and ED indicate the better naturalness of generated examples.



\begin{figure}[!tp]
    \centering
    \includegraphics[width=0.85\linewidth,scale=1]{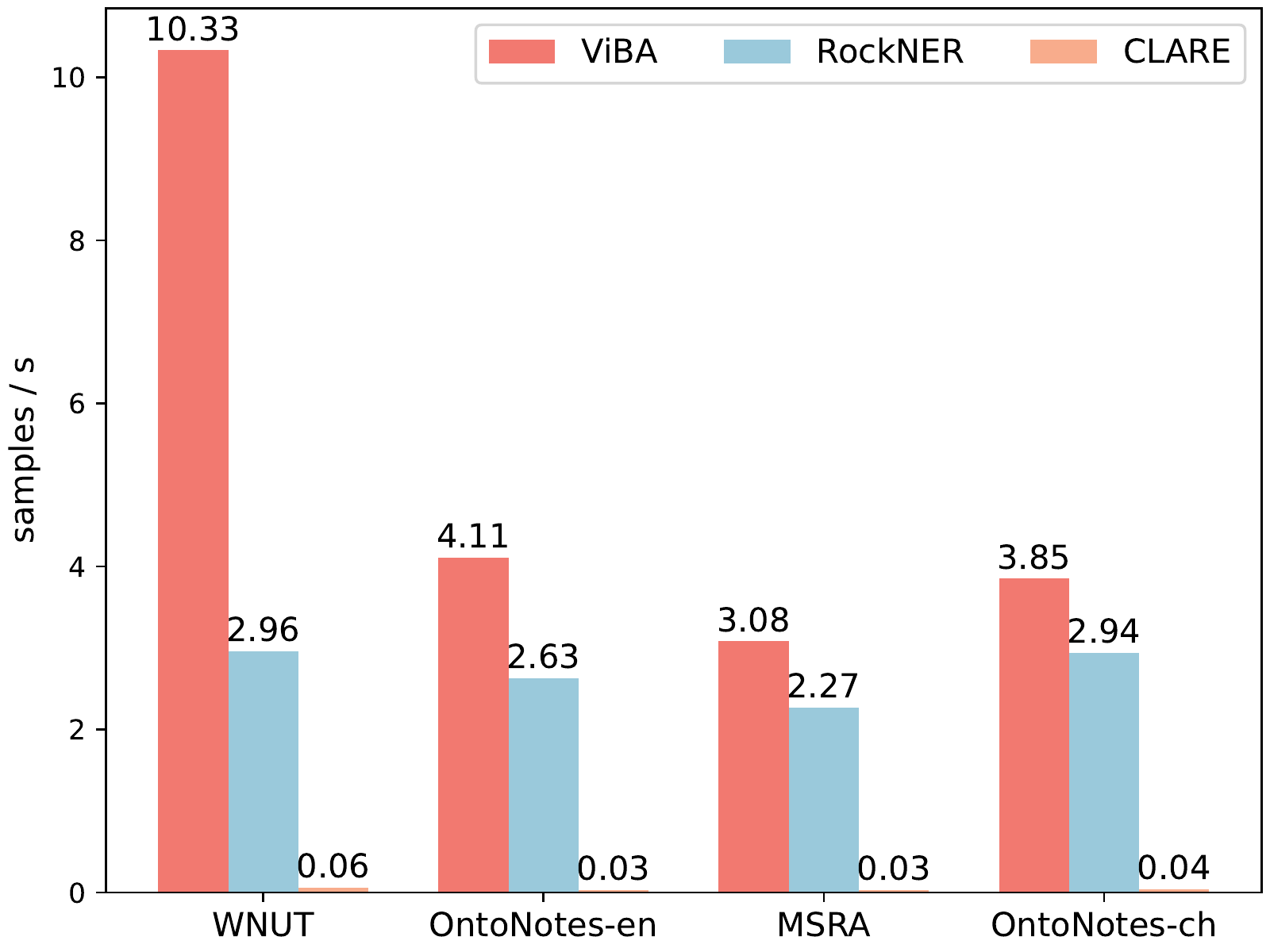}
    \caption{Comparison of time cost.}
    \label{fig:time_com}
\end{figure}

\subsection{Performance Comparison}

\subsection{Time Analysis}
The time complexity for ViBA to attack a sentence of length $n$ is about $O(m\times n)$, where $m$ is the amount of the named entities in this sentence. Usually, $m$ is much smaller than $n$. Thus, the time complexity is almost linear with $n$, which makes ViBA efficient. To verify it, we evaluate the number of samples that can be processed by ViBA, RockNER and CLARE within one second on the four datasets, as shown in Figure \ref{fig:time_com}. The victim model is RoBERTa-large.







Compared to the CLARE which runs the inference process of the pre-trained language model at least $4 n$ times for a sentence of length $n$, the attack speed of ViBA exceeds it more than 90 times. On the OntoNotes-en, MSRA and OntoNotes-ch, the speed of ViBA is 30\% - 60\% higher than RockNER. On the WNUT which has the average least entities and shortest sentences, it has surprising 250\% speed increase relative to RockNER. Moreover, ViBA also omits sophisticated preprocessing processes, such as entity linking, generating a huge candidate dictionary required by RockNER, which are extremely time expensive.

\section{Interpretation} \label{sec:interpretation}
This section interprets the effectiveness of ViBA and our motivation by empirical experiments.

\subsection{Boundary as Trigger} \label{sec:boundary_analysis}





As mentioned in \cite{lin-etal-2021-rockner}, the NER models tend to memorize the entity pattern instead of recognizing them by context-based reasoning, which prompts us to explore whether the boundary tokens are the memorized pattern for recognition. To this end, we mask out the boundary or inner words of an entity to expose which one causes the models to extract the entities more inaccurately.

\begin{figure}[!t]
    \centering
    \includegraphics[width=0.8\linewidth,scale=3]{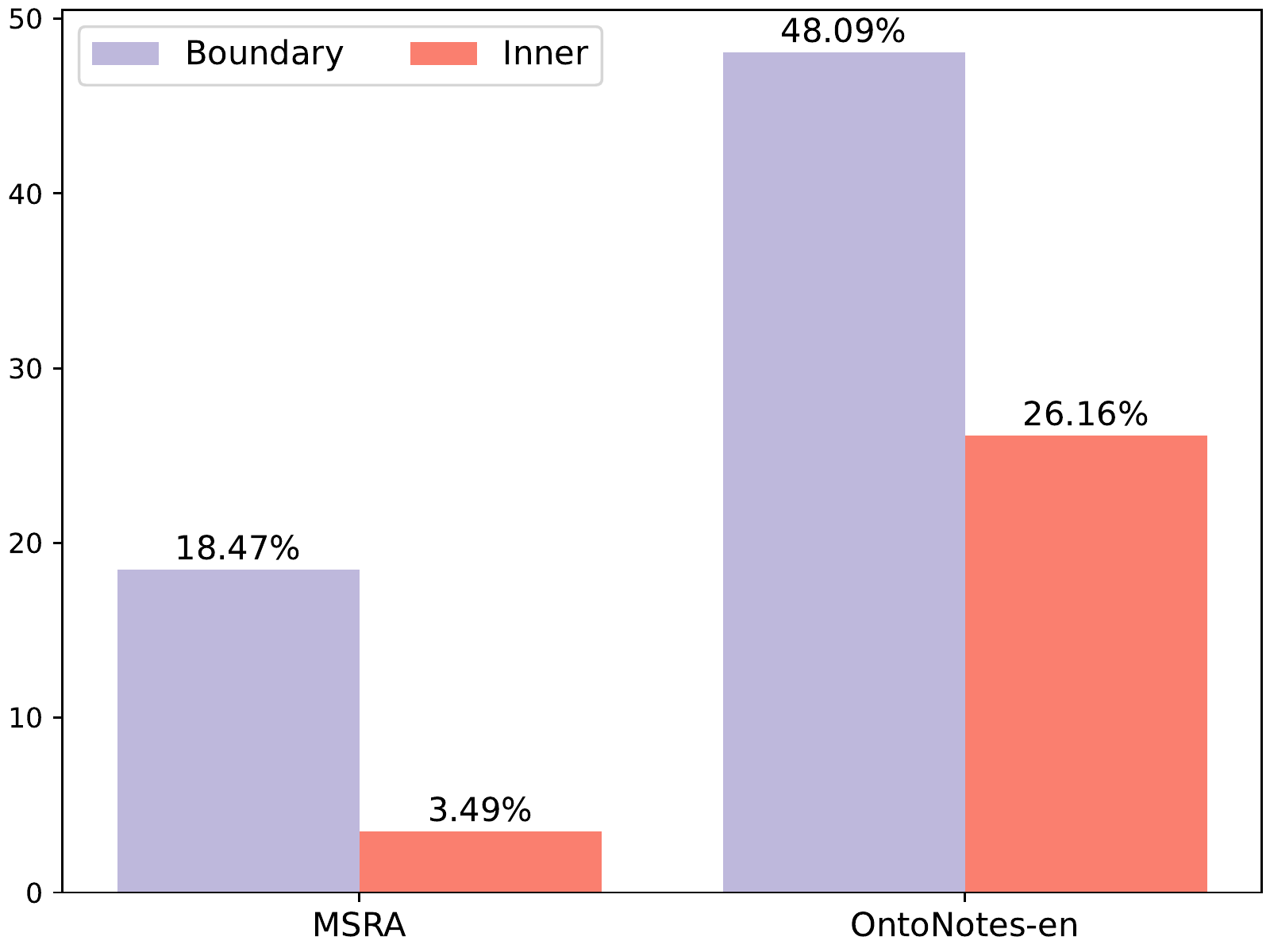}
    \caption{The probability drops caused by masking out boundary and inner tokens.}
    \label{fig:boundary_as_trigger}
\end{figure}

Specifically, we fine-tune two RoBERTa-large models on MSRA and OntoNotes-en datasets. Then we examine the models' dependence on the boundary and inner tokens: (1) For each sentence $\mathcal X=\mathbf{x}_1,\mathbf{x}_2,\cdots,\mathbf{x}_n$, one of its entities $e=\mathbf{x}_i,\mathbf{x}_{i+1},\cdots,\mathbf{x}_{i+m}$ is first recognized as type $t$ with the highest probability $p_t$ among all the types. (2) We mask out the boundary tokens $\mathbf{x}_i$ and $\mathbf{x}_{i+m}$ of $e$ in $\mathcal X$ respectively to obtain two sentences and feed them into the model again. The model separately estimates the probabilities $p_t^{'}$ and $p_t^{''}$ that the masked entities remain type $t$. Since $p_t^{'}$ and $p_t^{''}$ are always less than $p_t$, we leverage the mean value of two probability drops $p_t-p_t^{'}$, $p_t-p_t^{''}$ to reflect the dependence of the model on boundary. (3) Similarly, we mask out the inner tokens of $e$ and calculate the mean value of probability drops, as shown in Figure \ref{fig:boundary_as_trigger}.



\begin{figure}[!t]
    \centering
    \includegraphics[width=0.8\linewidth,scale=3]{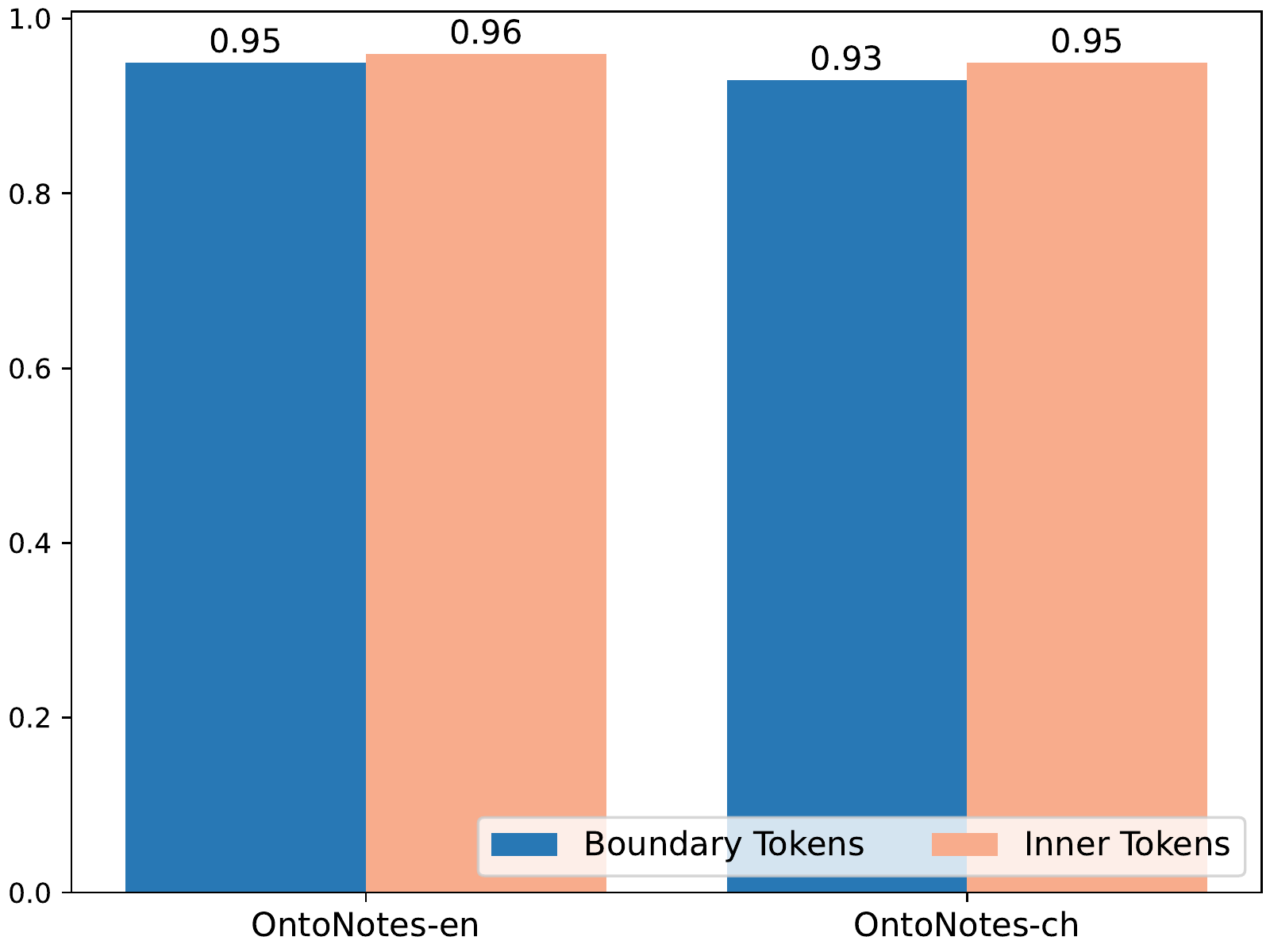}
    \caption{The cosine similarity of the hidden-states.}
    \label{fig:cosine_sim}
\end{figure}

On the Chinese MSRA dataset, the probability drop caused by masking out boundary tokens is more than five times that of masking out inner tokens. On English OntoNotes, masking out boundary tokens even causes a probability drop of nearly 50\%. It can be concluded that compared to masking out inner tokens, masking out the boundary tokens will significantly drop the probability that the model maintains the original judgment, which indicates that models may memorize boundary pattern and mainly rely on it to extract named entities. Therefore, we assume that inserting some boundary tokens into sentences can easily trigger the model to misidentify entities or wrongly recall entities by the memory effect as the occurrence of EBI, and it is our initial motivation.

\subsection{Robustness of Encoder and Decoder} \label{sec:robust_analysis}

The BERT-style NER models can be summarized into an encoder-decoder structure. The encoder usually leverages a strong pre-trained language model which encodes the input into contextual hidden-states. The decoder is usually served by the models such as MLP classifier, conditional random field (CRF), etc and classifies each token into a pre-defined label according to its hidden-states.





Since the hidden-states are the only medium between the encoder and decoder, we analyze their robustness from the stability of the hidden-states to further interpret ViBA. For each generated adversarial example $\mathfrak X$, it is fed into the encoder to obtain its hidden-states $\mathcal{H}$. Then we mask out the original entity in $\mathfrak X$ to get $\mathfrak X_{m}$ and input it into the encoder to obtain hidden-states $\mathcal{H}_{m}$. We select the representations of the inserted boundary from the $\mathcal{H},\mathcal{H}_{m}$ and calculate the cosine similarity between them. Similarly, we also calculate the cosine similarity for other tokens in the sentence. We conduct experiments with BERT-base on OntoNotes dataset. The average values of the cosine similarities are displayed in Figure \ref{fig:cosine_sim}.

From the results, we figure out that for the inserted boundary tokens, the cosine similarity of the hidden-states between the $\mathcal{H}$ and $\mathcal{H}_{m}$ exceeds 0.93 in two datasets. It is worth noting that the hidden-states of BERT-base are as high as 768 dimensions, and the cosine similarity so close to 1 shows that the inserted boundary does not cause a significant deviation of the encoder output. Similar to this phenomenon, other tokens also obtain an average similarity of 0.95 in two datasets, which further verifies that the encoder is relatively stable to $\mathfrak X$ and $\mathfrak X_{m}$. It can be concluded that when the hidden-states output by the encoder changed slightly in the position of the inserted boundary, the prediction of this boundary by the decoder will be confused. We summarize that for such an encoder-decoder NER model, our ViBA mainly attacks the non-robustness of the decoder.

To summary up, (1) The NER models tend to recognize the entities depending on the boundary, and perhaps memorize the boundary pattern. (2) The decoder is not robust enough to resist slight perturbation on hidden-states.







\begin{table}[!t]
\centering
{\begin{tabular}{lcc|cc}
\toprule
& \multicolumn{2}{c|}{\textbf{OntoNotes-en}} & \multicolumn{2}{c}{\textbf{OntoNotes-ch}}\\ 

              & \textbf{ASR}                   & \textbf{F$_1$}                 & \textbf{ASR}                 & \textbf{F$_1$}                     \\ \hline 
\begin{tabular}[c]{@{}l@{}} \textbf{FreeLB} \end{tabular}  & 70.5 & \textbf{89.5} & 86.0 & 85.2 \\ \hline
\begin{tabular}[c]{@{}l@{}} \textbf{ASA} \end{tabular}     & 72.2 & 89.3 & 86.8 & \textbf{85.3} \\ \hline 
\begin{tabular}[c]{@{}l@{}} \textbf{Mixed} \end{tabular}   & 68.6 & 75.4 & 77.7 & 84.1 \\ \hline \hline
 \textbf{$p$} & \textbf{ASR}                   & \textbf{F$_1$}                 & \textbf{ASR}                 & \textbf{F$_1$}                     \\ \hline

\begin{tabular}[c]{@{}l@{}} \textbf{0} \end{tabular}   & 73.2 & 89.2 & 85.5 & 85.0 \\ 
\hline
\begin{tabular}[c]{@{}l@{}}\textbf{0.3} \end{tabular}  & 63.7 & 88.8 & 87.1 & 84.7 \\ 
\hline
\begin{tabular}[c]{@{}l@{}}\textbf{0.5} \end{tabular}  & 67.7 & 88.3 & 85.4 & 83.6 \\ 
\hline
\begin{tabular}[c]{@{}l@{}}\textbf{0.8} \end{tabular}  & 69.8 & 83.1 & 71.5 & 63.0 \\ 

\bottomrule
\end{tabular}}
\caption{The results of masking out the boundary tokens for the encoder.}
\label{table:mask_boundary}
\end{table}

\begin{table}[!t]
\centering
{\begin{tabular}{lcc|cc}
\toprule
& \multicolumn{2}{c|}{\textbf{OntoNotes-en}} & \multicolumn{2}{c}{\textbf{OntoNotes-ch}}\\ 

\textbf{ } & \textbf{ASR}                   & \textbf{F$_1$}                 & \textbf{ASR}                 & \textbf{F$_1$}                     \\ \hline 
\begin{tabular}[c]{@{}l@{}} \textbf{WP} \end{tabular}  & 70.4 & 88.4 & 88.4 & 84.7 \\ \hline \hline
\textbf{$p$} & \textbf{ASR}                 & \textbf{F$_1$}                 & \textbf{ASR}                 & \textbf{F$_1$}                     \\ \hline
\begin{tabular}[c]{@{}l@{}} \textbf{0}  \end{tabular}  & 73.2 & 89.2 & 85.5 & 85.0 \\ 
\hline
\begin{tabular}[c]{@{}l@{}} \textbf{0.3} \end{tabular} & 70.2 & 88.8 & 85.7 & 85.1 \\ 
\hline
\begin{tabular}[c]{@{}l@{}}\textbf{0.5} \end{tabular}  & 70.8 & 88.7 & 84.7 & 85.0  \\ 
\hline
\begin{tabular}[c]{@{}l@{}}\textbf{0.8} \end{tabular}  & 75.1 & 87.6 & 80.4 & 84.3 \\ 

\bottomrule
\end{tabular}}
\caption{The results of applying the dropout to the hidden-states for the decoder and the weight perturbation baseline. }
\label{table:dropout_hidden}
\end{table}

\section{Defense Strategy: Boundary Cut}
This section presents a Boundary Cut strategy that helps the model resist ViBA from two aspects.


\subsection{Decouple Boundary and Inner Words} \label{sec:mask_Boundary}

Since the NER model recognizes the named entity relying more on the boundary pattern, a very straightforward idea is to the decouple boundary and inner words, and enhance the model to capture the pattern of inner words. We achieve this goal with the simplest way of masking out the boundary words at the input when training. In detail, we randomly mask out the left and right boundary tokens of an entity with a probability $p$ during the fine-tuning phase. Then we evaluate ViBA towards this model on the clean test set. In addition, to explore whether masking out the boundary words during training has an impact on the model's ability of recognizing the named entities, we also evaluate the $F_1$ on the test set, where a higher $F_1$ indicates a higher recognition performance. We apply BERT-base to conduct experiments on the OntoNotes dataset in Table \ref{table:mask_boundary}.

Compared to the case without masking ($p=0$), almost all ASR has a significant decrease after masking out the boundary words, which shows that the dosage of masking out boundary words is beneficial for resisting ViBA. An exception happens when $p=0.3$ on OntoNotes-ch which makes the model more fragile. Our explanation for this anomaly is that masking out the boundary words will cause a trade-off. On the one hand, it can reduce the model sensitivity to the boundary by decoupling information of the boundary and inner words, thus decreasing the ASR. On the other hand, it will also bring noise, which may lead to insufficient training and makes the model fragile. In this case, the latter may outweigh the former. When observing the recognition performance, the F$_1$ of all experiments slightly decreases as $p=0.3,0.5$ which indicates that the noise introduced by masking out the boundary does not cause much performance reduction. It is also not surprising that there is a large drop in F$_1$ with such big noise when $p=0.8$. Overall, when $p$ is within a reasonable range, masking out boundary can effectively resist ViBA without significantly reducing the recognition performance. Based on our experiments, $p=0.5$ works best.

Adversarial Training (AT) is the commonly used method to improve the model's robustness. We select FreeLB \cite{zhu2019freelb} and ASA \cite{DBLP:journals/corr/abs-2206-12608} as our baselines. Compared to them, although our F$_1$ is relatively lower, we have a significantly more advantageous ASR. Also, we retrain the model on the mixture of adversarial and original examples (Mixed), where we set the label of the inserted boundary to ``O'' in an adversarial example, and the rest of the tokens are consistent with the original example. To our surprise, Mixed significantly reduces ASR and does not damage F$_1$ substantially, especially for the Chinese dataset, which indicates the distinction between generated adversarial and original examples is really slight.

\subsection{Dropout Hidden-States}
Since the decoder is relatively non-robust to the hidden-states and ViBA mainly fools it, improving its robustness is also a direct idea. We propose to apply dropout \cite{hinton2012improving} on the hidden-states for enhancement. While also considering that the NER model is sensitive to boundary words, we randomly dropout the boundary of an entity on top of the hidden-states with a probability $p$. 
We conduct experiments on the OntoNotes dataset. The victim model is BERT-base with a vanilla MLP decoder. We take a classic weight perturbation (WP) method \cite{DBLP:conf/iclr/WenVBTG18} which can improve model robustness as the baseline.

As shown in Table \ref{table:dropout_hidden}, ASR drops significantly when $p=0.5,0.3$, meanwhile the F$_1$ on the test set is almost unaffected. ViBA also outperforms WP with a lower ASR and higher F$_1$. We can conclude that such a concise dropout can help the victim model resist ViBA without affecting its recognition performance. Also, the model is fragile due to the undertraining problem, and it is understandable to have poor ASR and F$_1$ when $p=0.8$.










\section{Related Work}

Current studies on robustness concentrate on text classification, question answering (QA), etc. For instance, \citeauthor{gao2018black} propose the DeepWordBug which effectively fools the models in a black-box scenario. SCPNs \cite{iyyer-etal-2018-adversarial} generates adversarial examples based on syntactic information for text classification. Popular TEXTFOOLER \cite{jin2020bert} attacks the BERT-style models with excellent ability and efficiency. BAE \cite{garg2020bae} is a black box attacker aiming at text classification and generates adversarial examples by contextual perturbations. CLARE \cite{li2021contextualized} produces fluent and grammatical examples through a mask-then-infill procedure. \citeauthor{gan-ng-2019-improving} attacks the question paraphrasing in the QA dataset. \citeauthor{tan-etal-2020-morphin} perturb the inflectional morphology of words to generate plausible and semantically similar adversarial examples. However, none of them aim at NER.

Recently, some researchers begin to focus on the robustness of NER models. \citeauthor{mayhew2020robust} study the impact of capitalization in NER models. \citeauthor{das2022resilience} explore the influence of the surrounding context perturbation on the entity. But none of them propose an efficient NER attacker. Nowadays, there are only a few studies that propose attackers for NER. Although Seqattack \cite{simoncini2021seqattack} adapts some above-mentioned attack methods for text classification to NER, it does not propose a new approach. RockNer \cite{lin-etal-2021-rockner} and Breaking BERT \cite{dirkson2021breaking} are rare NER attackers. But essentially, they bring up label shift issue and suffer from low efficiency as well as poor success rate.

\section{Conclusion}
This paper targets to study the robustness of current dominant NER models. Due to the label shift problem, existing attackers easily generate invalid adversarial examples. This paper first finds out the non-robust Entity Boundary Interference that is specific to NER and subsequently proposes an effective one-word modification attacker that alleviates label shift. Moreover, we interpret the effectiveness of our ViBA attacker and further propose a boundary cut strategy that can help the model defend against ViBA.







\bibliography{anthology,custom}

\begin{thebibliography}{31}
\expandafter\ifx\csname natexlab\endcsname\relax\def\natexlab#1{#1}\fi

\bibitem[{Babych and Hartley(2003)}]{babych2003improving}
Bogdan Babych and Anthony Hartley. 2003.
\newblock \href {https://aclanthology.org/W03-2201} {Improving machine
  translation quality with automatic named entity recognition}.
\newblock In \emph{Proceedings of the 7th International {EAMT} workshop on {MT}
  and other language technology tools, Improving {MT} through other language
  technology tools, Resource and tools for building {MT} at {EACL} 2003}.

\bibitem[{Clark et~al.(2018)Clark, Ji, and Smith}]{clark2018neural}
Elizabeth Clark, Yangfeng Ji, and Noah~A. Smith. 2018.
\newblock \href {https://doi.org/10.18653/v1/N18-1204} {Neural text generation
  in stories using entity representations as context}.
\newblock In \emph{Proceedings of the 2018 Conference of the North {A}merican
  Chapter of the Association for Computational Linguistics: Human Language
  Technologies, Volume 1 (Long Papers)}, pages 2250--2260, New Orleans,
  Louisiana. Association for Computational Linguistics.

\bibitem[{Cui et~al.(2020)Cui, Che, Liu, Qin, Wang, and
  Hu}]{cui-etal-2020-revisiting}
Yiming Cui, Wanxiang Che, Ting Liu, Bing Qin, Shijin Wang, and Guoping Hu.
  2020.
\newblock \href {https://doi.org/10.18653/v1/2020.findings-emnlp.58}
  {Revisiting pre-trained models for {C}hinese natural language processing}.
\newblock In \emph{Findings of the Association for Computational Linguistics:
  EMNLP 2020}, pages 657--668, Online. Association for Computational
  Linguistics.

\bibitem[{Das and Paik(2022)}]{das2022resilience}
Sudeshna Das and Jiaul Paik. 2022.
\newblock Resilience of named entity recognition models under adversarial
  attack.
\newblock In \emph{Proceedings of the First Workshop on Dynamic Adversarial
  Data Collection}, pages 1--6.

\bibitem[{Derczynski et~al.(2017)Derczynski, Nichols, van Erp, and
  Limsopatham}]{derczynski-etal-2017-results}
Leon Derczynski, Eric Nichols, Marieke van Erp, and Nut Limsopatham. 2017.
\newblock \href {https://doi.org/10.18653/v1/W17-4418} {Results of the
  {WNUT}2017 shared task on novel and emerging entity recognition}.
\newblock In \emph{Proceedings of the 3rd Workshop on Noisy User-generated
  Text}, pages 140--147, Copenhagen, Denmark. Association for Computational
  Linguistics.

\bibitem[{Devlin et~al.(2019)Devlin, Chang, Lee, and
  Toutanova}]{devlin2018bert}
Jacob Devlin, Ming-Wei Chang, Kenton Lee, and Kristina Toutanova. 2019.
\newblock \href {https://doi.org/10.18653/v1/N19-1423} {{BERT}: Pre-training of
  deep bidirectional transformers for language understanding}.
\newblock In \emph{Proceedings of the 2019 Conference of the North {A}merican
  Chapter of the Association for Computational Linguistics: Human Language
  Technologies, Volume 1 (Long and Short Papers)}, pages 4171--4186,
  Minneapolis, Minnesota. Association for Computational Linguistics.

\bibitem[{Dirkson et~al.(2021)Dirkson, Verberne, and
  Kraaij}]{dirkson2021breaking}
Anne Dirkson, Suzan Verberne, and Wessel Kraaij. 2021.
\newblock \href {https://arxiv.org/abs/2109.11308} {Breaking bert:
  Understanding its vulnerabilities for biomedical named entity recognition
  through adversarial attack}.
\newblock \emph{ArXiv preprint}, abs/2109.11308.

\bibitem[{Gan and Ng(2019)}]{gan-ng-2019-improving}
Wee~Chung Gan and Hwee~Tou Ng. 2019.
\newblock \href {https://doi.org/10.18653/v1/P19-1610} {Improving the
  robustness of question answering systems to question paraphrasing}.
\newblock In \emph{Proceedings of the 57th Annual Meeting of the Association
  for Computational Linguistics}, pages 6065--6075, Florence, Italy.
  Association for Computational Linguistics.

\bibitem[{Gao et~al.(2018)Gao, Lanchantin, Soffa, and Qi}]{gao2018black}
Ji~Gao, Jack Lanchantin, Mary~Lou Soffa, and Yanjun Qi. 2018.
\newblock Black-box generation of adversarial text sequences to evade deep
  learning classifiers.
\newblock In \emph{2018 IEEE Security and Privacy Workshops (SPW)}, pages
  50--56. IEEE.

\bibitem[{Garg and Ramakrishnan(2020)}]{garg2020bae}
Siddhant Garg and Goutham Ramakrishnan. 2020.
\newblock \href {https://doi.org/10.18653/v1/2020.emnlp-main.498} {{BAE}:
  {BERT}-based adversarial examples for text classification}.
\newblock In \emph{Proceedings of the 2020 Conference on Empirical Methods in
  Natural Language Processing (EMNLP)}, pages 6174--6181, Online. Association
  for Computational Linguistics.

\bibitem[{He et~al.(2020)He, Liu, Gao, and Chen}]{he2020deberta}
Pengcheng He, Xiaodong Liu, Jianfeng Gao, and Weizhu Chen. 2020.
\newblock \href {https://arxiv.org/abs/2006.03654} {Deberta: Decoding-enhanced
  bert with disentangled attention}.
\newblock \emph{ArXiv preprint}, abs/2006.03654.

\bibitem[{Hinton et~al.(2012)Hinton, Srivastava, Krizhevsky, Sutskever, and
  Salakhutdinov}]{hinton2012improving}
Geoffrey~E Hinton, Nitish Srivastava, Alex Krizhevsky, Ilya Sutskever, and
  Ruslan~R Salakhutdinov. 2012.
\newblock \href {https://arxiv.org/abs/1207.0580} {Improving neural networks by
  preventing co-adaptation of feature detectors}.
\newblock \emph{ArXiv preprint}, abs/1207.0580.

\bibitem[{Iyyer et~al.(2018)Iyyer, Wieting, Gimpel, and
  Zettlemoyer}]{iyyer-etal-2018-adversarial}
Mohit Iyyer, John Wieting, Kevin Gimpel, and Luke Zettlemoyer. 2018.
\newblock \href {https://doi.org/10.18653/v1/N18-1170} {Adversarial example
  generation with syntactically controlled paraphrase networks}.
\newblock In \emph{Proceedings of the 2018 Conference of the North {A}merican
  Chapter of the Association for Computational Linguistics: Human Language
  Technologies, Volume 1 (Long Papers)}, pages 1875--1885, New Orleans,
  Louisiana. Association for Computational Linguistics.

\bibitem[{Jin et~al.(2020)Jin, Jin, Zhou, and Szolovits}]{jin2020bert}
Di~Jin, Zhijing Jin, Joey~Tianyi Zhou, and Peter Szolovits. 2020.
\newblock \href {https://aaai.org/ojs/index.php/AAAI/article/view/6311} {Is
  {BERT} really robust? {A} strong baseline for natural language attack on text
  classification and entailment}.
\newblock In \emph{The Thirty-Fourth {AAAI} Conference on Artificial
  Intelligence, {AAAI} 2020, The Thirty-Second Innovative Applications of
  Artificial Intelligence Conference, {IAAI} 2020, The Tenth {AAAI} Symposium
  on Educational Advances in Artificial Intelligence, {EAAI} 2020, New York,
  NY, USA, February 7-12, 2020}, pages 8018--8025. {AAAI} Press.

\bibitem[{Levow(2006)}]{levow2006third}
Gina-Anne Levow. 2006.
\newblock \href {https://aclanthology.org/W06-0115} {The third international
  {C}hinese language processing bakeoff: Word segmentation and named entity
  recognition}.
\newblock In \emph{Proceedings of the Fifth {SIGHAN} Workshop on {C}hinese
  Language Processing}, pages 108--117, Sydney, Australia. Association for
  Computational Linguistics.

\bibitem[{Li et~al.(2021)Li, Zhang, Peng, Chen, Brockett, Sun, and
  Dolan}]{li2021contextualized}
Dianqi Li, Yizhe Zhang, Hao Peng, Liqun Chen, Chris Brockett, Ming-Ting Sun,
  and Bill Dolan. 2021.
\newblock \href {https://doi.org/10.18653/v1/2021.naacl-main.400}
  {Contextualized perturbation for textual adversarial attack}.
\newblock In \emph{Proceedings of the 2021 Conference of the North American
  Chapter of the Association for Computational Linguistics: Human Language
  Technologies}, pages 5053--5069, Online. Association for Computational
  Linguistics.

\bibitem[{Lin et~al.(2021)Lin, Gao, Yan, Moreno, and
  Ren}]{lin-etal-2021-rockner}
Bill~Yuchen Lin, Wenyang Gao, Jun Yan, Ryan Moreno, and Xiang Ren. 2021.
\newblock \href {https://doi.org/10.18653/v1/2021.emnlp-main.302} {{R}ock{NER}:
  A simple method to create adversarial examples for evaluating the robustness
  of named entity recognition models}.
\newblock In \emph{Proceedings of the 2021 Conference on Empirical Methods in
  Natural Language Processing}, pages 3728--3737, Online and Punta Cana,
  Dominican Republic. Association for Computational Linguistics.

\bibitem[{Liu et~al.(2019)Liu, Ott, Goyal, Du, Joshi, Chen, Levy, Lewis,
  Zettlemoyer, and Stoyanov}]{liu2019roberta}
Yinhan Liu, Myle Ott, Naman Goyal, Jingfei Du, Mandar Joshi, Danqi Chen, Omer
  Levy, Mike Lewis, Luke Zettlemoyer, and Veselin Stoyanov. 2019.
\newblock \href {https://arxiv.org/abs/1907.11692} {Roberta: A robustly
  optimized bert pretraining approach}.
\newblock \emph{ArXiv preprint}, abs/1907.11692.

\bibitem[{Mayhew et~al.(2020)Mayhew, Gupta, and Roth}]{mayhew2020robust}
Stephen Mayhew, Nitish Gupta, and Dan Roth. 2020.
\newblock \href {https://aaai.org/ojs/index.php/AAAI/article/view/6368} {Robust
  named entity recognition with truecasing pretraining}.
\newblock In \emph{The Thirty-Fourth {AAAI} Conference on Artificial
  Intelligence, {AAAI} 2020, The Thirty-Second Innovative Applications of
  Artificial Intelligence Conference, {IAAI} 2020, The Tenth {AAAI} Symposium
  on Educational Advances in Artificial Intelligence, {EAAI} 2020, New York,
  NY, USA, February 7-12, 2020}, pages 8480--8487. {AAAI} Press.

\bibitem[{Morris et~al.(2020)Morris, Lifland, Yoo, Grigsby, Jin, and
  Qi}]{morris2020textattack}
John Morris, Eli Lifland, Jin~Yong Yoo, Jake Grigsby, Di~Jin, and Yanjun Qi.
  2020.
\newblock \href {https://doi.org/10.18653/v1/2020.emnlp-demos.16}
  {{T}ext{A}ttack: A framework for adversarial attacks, data augmentation, and
  adversarial training in {NLP}}.
\newblock In \emph{Proceedings of the 2020 Conference on Empirical Methods in
  Natural Language Processing: System Demonstrations}, pages 119--126, Online.
  Association for Computational Linguistics.

\bibitem[{Nikoulina et~al.(2012)Nikoulina, Sandor, and
  Dymetman}]{nikoulina2012hybrid}
Vassilina Nikoulina, Agnes Sandor, and Marc Dymetman. 2012.
\newblock \href {https://aclanthology.org/W12-5701} {Hybrid adaptation of named
  entity recognition for statistical machine translation}.
\newblock In \emph{Proceedings of the Second Workshop on Applying Machine
  Learning Techniques to Optimise the Division of Labour in Hybrid {MT}}, pages
  1--16, Mumbai, India. The COLING 2012 Organizing Committee.

\bibitem[{Peng and Dredze(2016)}]{peng2016improving}
Nanyun Peng and Mark Dredze. 2016.
\newblock \href {https://doi.org/10.18653/v1/P16-2025} {Improving named entity
  recognition for {C}hinese social media with word segmentation representation
  learning}.
\newblock In \emph{Proceedings of the 54th Annual Meeting of the Association
  for Computational Linguistics (Volume 2: Short Papers)}, pages 149--155,
  Berlin, Germany. Association for Computational Linguistics.

\bibitem[{Sil and Yates(2013)}]{sil2013re}
Avirup Sil and Alexander Yates. 2013.
\newblock \href {https://doi.org/10.1145/2505515.2505601} {Re-ranking for joint
  named-entity recognition and linking}.
\newblock In \emph{22nd {ACM} International Conference on Information and
  Knowledge Management, CIKM'13, San Francisco, CA, USA, October 27 - November
  1, 2013}, pages 2369--2374. {ACM}.

\bibitem[{Simoncini and Spanakis(2021)}]{simoncini2021seqattack}
Walter Simoncini and Gerasimos Spanakis. 2021.
\newblock Seqattack: On adversarial attacks for named entity recognition.
\newblock In \emph{Proceedings of the 2021 Conference on Empirical Methods in
  Natural Language Processing: System Demonstrations}, pages 308--318.

\bibitem[{Tan et~al.(2020{\natexlab{a}})Tan, Qiu, Chen, Wang, and
  Huang}]{tan2020boundary}
Chuanqi Tan, Wei Qiu, Mosha Chen, Rui Wang, and Fei Huang. 2020{\natexlab{a}}.
\newblock Boundary enhanced neural span classification for nested named entity
  recognition.
\newblock In \emph{Proceedings of the AAAI Conference on Artificial
  Intelligence}, volume~34, pages 9016--9023.

\bibitem[{Tan et~al.(2020{\natexlab{b}})Tan, Joty, Kan, and
  Socher}]{tan-etal-2020-morphin}
Samson Tan, Shafiq Joty, Min-Yen Kan, and Richard Socher. 2020{\natexlab{b}}.
\newblock \href {https://doi.org/10.18653/v1/2020.acl-main.263} {It{'}s
  morphin{'} time! {C}ombating linguistic discrimination with inflectional
  perturbations}.
\newblock In \emph{Proceedings of the 58th Annual Meeting of the Association
  for Computational Linguistics}, pages 2920--2935, Online. Association for
  Computational Linguistics.

\bibitem[{Weischedel et~al.(2013)Weischedel, Palmer, Marcus, Hovy, Pradhan,
  Ramshaw, Xue, Taylor, Kaufman, Franchini et~al.}]{weischedel2013ontonotes}
R~Weischedel, M~Palmer, M~Marcus, E~Hovy, S~Pradhan, L~Ramshaw, N~Xue,
  A~Taylor, J~Kaufman, M~Franchini, et~al. 2013.
\newblock Ontonotes release 5.0 ldc2013t19. linguistic data consortium,
  philadelphia, pa (2013).

\bibitem[{Wen et~al.(2018)Wen, Vicol, Ba, Tran, and
  Grosse}]{DBLP:conf/iclr/WenVBTG18}
Yeming Wen, Paul Vicol, Jimmy Ba, Dustin Tran, and Roger~B. Grosse. 2018.
\newblock \href {https://openreview.net/forum?id=rJNpifWAb} {Flipout: Efficient
  pseudo-independent weight perturbations on mini-batches}.
\newblock In \emph{6th International Conference on Learning Representations,
  {ICLR} 2018, Vancouver, BC, Canada, April 30 - May 3, 2018, Conference Track
  Proceedings}. OpenReview.net.

\bibitem[{Wu and Zhao(2022)}]{DBLP:journals/corr/abs-2206-12608}
Hongqiu Wu and Hai Zhao. 2022.
\newblock \href {https://arxiv.org/abs/2206.12608} {Adversarial self-attention
  for language understanding}.
\newblock \emph{ArXiv preprint}, abs/2206.12608.

\bibitem[{Xu(2022)}]{Text2vec}
Ming Xu. 2022.
\newblock Text2vec: Text to vector toolkit.
\newblock \url{https://github.com/shibing624/text2vec}.

\bibitem[{Zhu et~al.(2020)Zhu, Cheng, Gan, Sun, Goldstein, and
  Liu}]{zhu2019freelb}
Chen Zhu, Yu~Cheng, Zhe Gan, Siqi Sun, Tom Goldstein, and Jingjing Liu. 2020.
\newblock \href {https://openreview.net/forum?id=BygzbyHFvB} {Freelb: Enhanced
  adversarial training for natural language understanding}.
\newblock In \emph{8th International Conference on Learning Representations,
  {ICLR} 2020, Addis Ababa, Ethiopia, April 26-30, 2020}. OpenReview.net.

\end{thebibliography}
\bibliographystyle{acl_natbib}




\end{document}